\documentclass[conference]{IEEEtran}
\IEEEoverridecommandlockouts
\usepackage{cite}
\usepackage{amsmath,amssymb,amsfonts}
\usepackage{graphicx}
\usepackage{textcomp}
\usepackage{xcolor}
\usepackage{algpseudocode}
\usepackage{algorithm}
\usepackage{multicol}
\usepackage{multirow}
\usepackage{amssymb}
\usepackage{booktabs}
\usepackage{marvosym}
\def\BibTeX{{\rm B\kern-.05em{\sc i\kern-.025em b}\kern-.08em
    T\kern-.1667em\lower.7ex\hbox{E}\kern-.125emX}}
\begin{document}

\title{SOLE: Hardware-Software Co-design of \underline{So}ftmax and \underline{L}ayerNorm for \underline{E}fficient Transformer Inference\\
\thanks{Yongpan Liu and Peiqin Sun are the corresponding authors.}
}
\author{\IEEEauthorblockN{ Wenxun Wang\IEEEauthorrefmark{1},
                            Shuchang Zhou\IEEEauthorrefmark{2},
                            Wenyu Sun\IEEEauthorrefmark{1},
                            Peiqin Sun\IEEEauthorrefmark{2} and 
                            Yongpan Liu\IEEEauthorrefmark{1}}
\IEEEauthorblockA{\IEEEauthorrefmark{1} Department of Electronic Engineering, Tsinghua University, Beijing, China}
\IEEEauthorblockA{\IEEEauthorrefmark{2} MEGVII Technology, Beijing, China}
\IEEEauthorblockA{wx-wang23@mails.tsinghua.edu.cn, wy-sun@sz.tsinghua.edu.cn, ypliu@tsinghua.edu.cn, \IEEEauthorrefmark{2}\{zsc,sunpeiqin\}@megvii.com}}


\maketitle

\begin{abstract}
Transformers have shown remarkable performance in both natural language processing (NLP) and computer vision (CV) tasks. However, their real-time inference speed and efficiency are limited due to the inefficiency in Softmax and Layer Normalization (LayerNorm). Previous works based on function approximation suffer from inefficient implementation as they place emphasis on computation while disregarding memory overhead concerns. Moreover, such methods rely on retraining to compensate for approximation error which can be costly and inconvenient.

In this paper, we present SOLE, a hardware-software co-design for Softmax and LayerNorm which is composed of E2Softmax and AILayerNorm. E2Softmax utilizes log2 quantization of exponent function and log-based division to approximate Softmax while AILayerNorm adopts low-precision statistic calculation. Compared with state-of-the-art designs, we achieve both low-precision calculation and low bit-width storage on Softmax and LayerNorm. Experiments show that SOLE maintains inference accuracy without retraining while offering orders of magnitude speedup and energy savings over GPU, achieving 3.04$\times$, 3.86$\times$ energy-efficiency improvements and 2.82$\times$, 3.32$\times$ area-efficiency improvements over prior state-of-the-art custom hardware for Softmax and LayerNorm, respectively.
\end{abstract}


\section{Introduction}
\begin{figure}[!t]
\includegraphics[width=0.5\textwidth,height=0.33\textwidth]{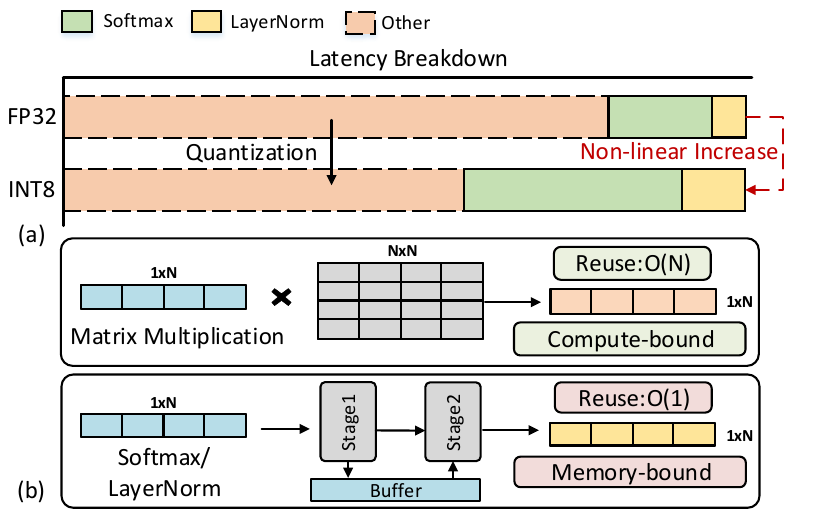}
  \caption{(a) Latency Breakdown of Deit-Tiny with the image size of 448 on a 2080Ti GPU. (b) Difference between MatMul operations with Softmax/LayerNorm operations.}
  \label{fig:1}
\end{figure}

In recent years, transformer-based networks\cite{Transformer} have achieved success in enhancing the performance of computer vision (CV) tasks\cite{cv-transformer, cv2, cv3} as well as natural language processing (NLP) tasks\cite{nlp-transformer, nlp2, nlp3}. Despite their impressive performance, their computational characteristics have become a non-negligible defect. Compared to other networks, like CNNs \cite{CNN} and RNNs \cite{rnn},  transformers have a large number of parameters and massive computation overhead, i.e. ViT-22B\cite{ViT-22B}, GPT-3\cite{GPT3}, PaLM\cite{palm}. Efforts have been made to mitigate the problem. Algorithms like quantization \cite{ptq,dorefa,q-bert} are developed to diminish memory footprint and reduce computation overhead while many accelerators have also been proposed to expedite inference \cite{DOTA,SpAtten,Sanger,a3}. These endeavors primarily focused on accelerating the matrix multiplication in transformers, which is widely considered the primary bottleneck of conventional network, such as CNNs and MLPs.
However, little attention has been paid to the non-linear computations that exist extensively in self-attention mechanism, including Softmax and Layer Normalization (LayerNorm). Implemented by 32-bit floating-point (FP32) arithmetic units, these operations severely restrict the enhancement of end-to-end transformer inference as reported in \cite{Softermax}. As shown in Fig. \ref{fig:1}(a), this issue becomes worse when matrix multiplications are calculated using 8-bit integer (INT8) arithmetic after quantization, emphasizing the urgent need for a solution.

Previous works have proposed methods based on function approximation to optimize these operations. However, the effectiveness of these works is limited to computation, and storage concerns are often overlooked. Fig. \ref{fig:1}(b) illustrates two distinct differences between Softmax/LayerNorm operations and traditional matrix multiplication. Firstly, Softmax/LayerNorm operations employ a two-stage dataflow, which requires buffering intermediate data since the output cannot be generated directly. Secondly, due to the rarity of input data reuse in these operations, it is challenging to amortize memory costs through computation. Collectively, these two factors contribute to the memory-bound issue of Softmax/LayerNorm operations. Prior works including Softermax\cite{Softermax} and I-BERT \cite{I-BERT} fail to address this problem as they still need to buffer 16-bit and 32-bit data in the process, respectively. Moreover, retraining or fine-tuning is usually required for such works as compensation of approximation errors, incurring additional training overhead that becomes increasingly expensive as transformers grow larger.

In this paper, we propose SOLE, a hardware/software co-design method to solve these problems. In SOLE, we design E2Softmax and AILayerNorm for Softmax and LayerNorm, respectively. For Softmax, log2 quantization is appiled on the output of exponent function, which compresses intermediate data to 4-bit and significantly mitigates the memory-bound issue. With log2 quantization, the exponent function is implemented through a specialized Log2Exp Unit that is composed solely of shifters and adders. Besides, floating-point division is substituted with the proposed Approximate Log-based Division to fully utilize the log2-quantized output. In this way, E2Softmax can be implemented multiplication-free and LUT-free. For LayerNorm, we find that the statistic calculation is resilient to errors introduced by small variation in inputs. Hence, we adopt dynamic compression and combine it with Power-of-Two Factor (PTF) \cite{FQ-VIT}, uniformly optimizing the dataflow to achieve low-precision statistic calculation. Consequently, AILayerNorm requires only 8-bit data buffering and 4-bit multiplication for statistic calculation.

In summary, our contributions can be concluded as follows:

\begin{itemize} 
\item We propose Efficient log2 quantized Softmax (E2Softmax), a hardware-friendly softmax algorithm based on log2 quantization of exponent function. 
\item We propose Approximate Integer Layer Normalization (AILayerNorm), an efficient layernorm algorithm with low-precision statistic calculation.
\item Exclusive software experiments have been conducted on CV and NLP tasks with various transformer-based network. Results show that SOLE incurs negligible accuracy drops without additional training.
\item Hardware experiments demonstrate that SOLE delivers 36.2x and 61.3x average speedup, as well as orders-of-magnitude energy-efficiency improvements over GPU for Softmax and LayerNorm, respectively. In comparison to state-of-the-art custom hardware, SOLE provides 2.82x and 3.32x area-efficiency improvements and 3.04x and 3.86x energy-efficiency improvements for Softmax and LayerNorm, respectively
\end{itemize}
\begin{figure}[!t]
\includegraphics[width=0.5\textwidth,height=0.38\textwidth]{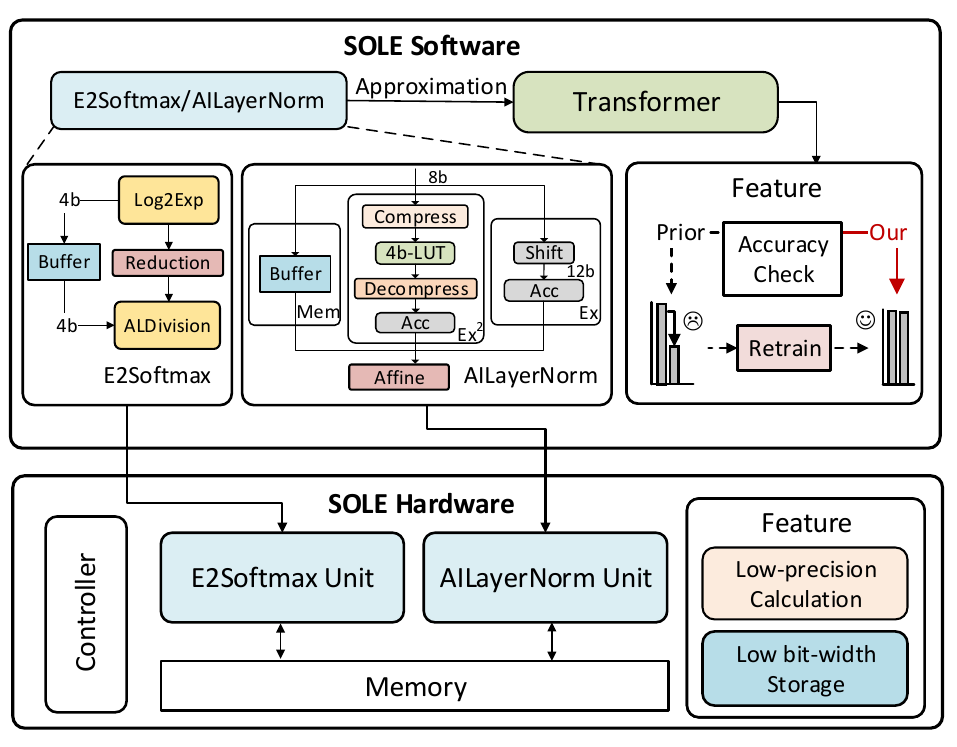}
  \centering
  \caption{Overview of SOLE}
  \label{fig:2}
\end{figure}

\section{Related work}
\subsection{Prominent Non-Linear Operations in Transformers}
A typical transformer model consists of stacked encoder and decoder blocks, each of which is composed of multiple layers of two sub-modules: a multi-head self-attention block and a feed-forward network\cite{Transformer}. Non-linear operations occupy significant parts in the transformer computation\cite{Softermax}, especially Softmax and LayerNorm. The Softmax function is used to compute the attention weights for the multi-head self-attention mechanism, while LayerNorm is used to normalize the hidden states at each layer of the model. Softmax and LayerNorm can be defined as follows:
\begin{equation}
    \begin{split}
        \mathrm{Softmax}(X_i) = \frac{\mathrm{exp}(X_i-X_{max})}{\sum_{j=1}^{n}\mathrm{exp}(X_j-X_{max})} \\
        \mathrm{LayerNorm}(X_i) = \frac{X_i-\mu}{\sigma} \cdot \gamma + \beta
    \end{split}
\end{equation}
where $\mu = \frac{1}{C} \sum_{i=1}^{C}X_i, \sigma = \sqrt{\frac{1}{C}\sum_{i=1}^{C}(X_i-\mu)^2}$. These operations contribute a large fraction of run-time in transformer inference when implemented with costly FP32 arithmetic computation\cite{Softermax}. For hardware targeting low-end device, it is considered as significant overhead \cite{survey}. Therefore, several methods have been proposed to address this issue.

\subsection{Accelerating Non-Linear Operations}
A conventional method for accelearating non-linear operations is function approximation \cite{sm_prior0,prior1}. For instance, Softermax \cite{Softermax} proposed low-precision computation for Softmax. I-BERT \cite{I-BERT} used 32-bit integer arithmetic as an approximation of Softmax, LayerNorm and other non-linear operations. NN-LUT \cite{NN-LUT} adpoted hardware-friendly Look-up table (LUT) and linear piece-wise approximation of non-linear operations based on I-BERT code base. The LUT contents were obtained through one-hidden-layer ReLU network. Desipte the benefits, there are two notable drawbacks associated with these approaches. Firstly, fine-tuning or retraining is required for prior works \cite{I-BERT,Softermax} to recover the performance, incurring increasing costs as transformers grow larger. Secondly, these works fail to make efficient implementation of Softmax and LayerNorm since both high-precision multiplication and high bit-width data storage are still indispensable. 

Fig. \ref{fig:2} shows the overview of SOLE. Unlike prior works, SOLE develops E2Softmax and AILayerNorm algorithms specifically designed for Softmax and LayerNorm. These algorithms incur minimal accuracy loss without the requirement for retraining. Additionally, SOLE introduces custom hardware units to support these algorithm, enabling efficient implementation through low-precision calculation and low bit-width storage.

\section{SOLE Software}
In SOLE, we design E2Softmax and AILayerNorm algorithms for efficient Softmax and LayerNorm implementation. In this section, we first introduce the preliminaries of our algorithms. Then, we explain the details of E2Softmax and AILayerNorm algorithms.
\subsection{Preliminary}
We explain the techniques relevant to the proposed algorithms in this section.

\noindent \textbf{Log2 Quantization.} Generally speaking, it converts continues value in $(0,1)$ into a set of discrete numbers like ${2^0,2^{-1},2^{-2}}$ \cite{log2}. Assuming the quantization bit-width $b$, the log2 quantization process can be defined as:
\begin{equation}
    \text{Log2Q}(X) = \text{Clip}(
    \lfloor - \mathrm{log}_2(X) \rceil , 0 , 2^{b} - 1
    ),X \in (0,1)
\end{equation}

\noindent \textbf{Log-based Division.} Proposed by Mitchell \cite{Mitchell}, it uses linear approximation to compute the logarithm of binary numbers and performs division by subtract and shift operations. Suppose an unsigned N-bit integer $X$, it can be defined as:
\begin{equation}
    X = \sum_{i=0}^{N-1} 2^ib_i = 2^{k_x} + \sum_{i=0}^{k_x-1} 2^ib_i=2^{k_x}(1+x)
\end{equation}
where $k_x$ represents the leading-one bit of $X$ and $x \in (0,1)$. The approximation of the logarithm is described as $\mathrm{log}_2(X) = k_x + x$, where $k_x$ is the characteristic part and $x$ is the decimal part.
When it comes to the division, for instance, $Q = \frac{X_1}{X_2}$, we can approximate the log of quotient as:
\begin{equation}
    \mathrm{log}_2(Q) = k_1 + x_1 - k_2 -x_2
\end{equation}

The approximate quotient is calculated by applying the inverse of the approximate logarithm:
\begin{equation}
    Q^{'} = \left\{
	\begin{aligned}
	2^{k_1-k_2-1}(2+x_1-x_2)&,\quad x_1 - x_2 < 0 \\
    2^{k_1-k_2}  (1+x_1-x_2)&,\quad x_1 - x_2 \geq 0
	\end{aligned}
	\right.
\end{equation}

Since the inverse of logarithm requires the fractional part is in $(0,1)$, the expression is divided into two parts in case a borrow is taken from the characteristic part when $x_1 - x_2 < 0$.

\noindent \textbf{Power-of-Two Factor (PTF).} Proposed by \cite{FQ-VIT} for LayerNorm quantization, PTF equips different channels with different factors to address the inter-channel variation in the inputs of LayerNorm layers. Assuming the input activation $\mathrm{X} \in \mathbb{R}^{\mathrm{B}\times \mathrm{L} \times \mathrm{C}}$, the quantization bit-width $b$, the layer-wise quantization parameters $\rm s,zp \in \mathbb{R}^1$,and the PTF $\alpha \in \mathbb{N}^{\mathrm{C}}$, the quantized activation $\mathrm{X_Q}$ can be defined as:
\begin{equation}
  X_Q = \text{Clip}( \lfloor \frac{X}{2^{\alpha}s} \rceil + \mathrm{zp},0,2^b - 1)
\end{equation}
where $s$ and $\rm zp$ stand for the quantization scale and zero point, respectively. Generally speaking, channels with large range of activation will have larger $\alpha$ such that their scaling factors appropriately match their range.
\subsection{E2Softmax}
\textrm{E2Softmax} algorithm adopts several techniques to improve efficiency. Firstly, we apply log2 quantization on the output of exponent function. Secondly, we substitute floating-point division and exponent function with approximate log-based division and Log2Exp function, as shown in Algorithm \ref{E2S}. All computations are carried out in fixed-point precision with 8-bit quantized inputs. An online normalization scheme \cite{sm_online} is added to further reduce latency.
\begin{figure}[!t]
\includegraphics[width=0.5\textwidth,height=0.27\textwidth]{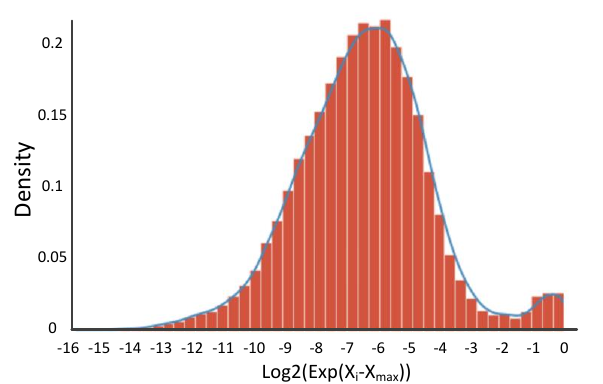}
  \caption{Distribution of $\mathbf{Exp}(X_i-X_{max})$ in logarithm. }
  \label{fig:log}
\end{figure}

\noindent \textbf{Log2 Quantization of Exponent Output.} The exponent function is a crucial component of Softmax, but its non-linearity makes it expensive to implement. Typically, multiplication and large LUTs are required to perform linear piece-wise approximation, inducing power and area burden. To alleviate the issue, we apply log2 quantization on the output of exponent operation so that it can be implemented hardware-friendly, which can be defined as follows:
\begin{equation}
    \textrm{Log2Exp}(x) = - \lfloor \textrm{log2}(e^{x}) \rceil = - \lfloor (x) \times \frac{1}{ln2} \rceil
\end{equation}
$\frac{1}{ln2}$ can be approximated as $1.4375$. Hence, the function is simply:
\begin{equation}
    \textrm{Log2Exp}(x) = - \lfloor x + x >> 1 - x >> 4 \rceil, x \in \left (-\infty,0 \right ]
\end{equation}

We adopt this transformation mainly for three reasons. Firstly, the inputs of exponent function need to perform subtraction with the maximum of the related vector to avoid overflow, naturally satisfying the input range of \textrm{Log2Exp} function. Secondly, the distribution of the exponent output is similar to the normalization distribution when plotted on a log2 scale (as shown in Fig. \ref{fig:log}), making the log2 method an ideal quantization choice. Thirdly, the Softmax function is concerned with the relative value  of the exponent output instead of the absolute value. Therefore, the error introduced through log2 quantization of the exponent output will be decreased after division \cite{SpAtten}. In this way, we can also implement exponent function in a more efficient way. Experiments in Table \ref{table1} and Table \ref{table2} show that the exponent output can be quantized to 4-bit with minimal accuracy drop through log2 quantization, resulting in further downsizing of memory footprints.


\noindent \textbf{Approximate Log-based Division.} Another non-linear operation in Softmax is division, which also relies large LUTs and high-precision multipliers. Since 4-bit logarithm of the exponent output is obtained through log2 quantization, it is reasonable to leverage log-based division to further reduce computational complexity. Assuming the log2 quantized exponent output $k_y = \mathrm{Log2Exp}(Q_y) $ and the reduced sum $S= 2 ^ {k_s} \cdot(1+s),s \in (0,1)$, we have:
\begin{equation}
    \begin{split}
\frac{Q_y}{S} = 2^{-(k_y + k_s)}\cdot\frac{1}{1+s}
    \end{split}
\end{equation}
$k_s$ represents the position of the leading-one of $S$ and $s$ stands for the rest bits. We adopt linear approximation and quantize the $s$ to 1-bit $q(s)$:
\begin{equation}
    \begin{split}
    q(s) &= \frac{\lfloor 2\cdot s \rfloor}{2} ,\ q(s) \in \lbrace 0,0.5 \rbrace \\
\frac{Q_y}{S} &\approx 2^{-(k_y + k_s + 1)} \cdot (1 - q(s)) 
    \end{split}
\end{equation}
Therefore, the error between approximate division and full-precision version can be formulated as:
\begin{equation}
    \delta = 2^{-(k_y + k_s)} \cdot \frac{s - 1 - q(s)\cdot (s+1)}{2(1+s)}
\end{equation}
Considering $s$ as an uniform random variable that takes value in $\left[ 0,1 \right]$, we can get the expectation of the error:
\begin{equation}
\begin{split}
    E(\delta) &= 2^{-(k_y+k_s)}(\int_0^{0.5} \frac{s-1}{2(s+1)}ds + \int_{0.5}^{1}\frac{s - 3}{4(1+s)}ds) \\
    &= 2^{-(k_y+k_s+1)} \times (\text{-}0.636)
\end{split}
\end{equation}
To make our approximation unbiased, we modify the Approximate Log-based Division (ALDivision) as:
\begin{equation}
    \begin{split}
        \mathrm{ALDivision}(k_y,S) = 2^{-(k_y + k_s + 1)} \cdot (1.636 - q(s))
    \end{split}
\end{equation}

In this way, we can perform division through hardware-friendly shift and subtraction operation. In fact, the subtraction can be replaced with a two-way multiplexer since $q(s)$ is a 1-bit number. Implementation details will be discussed in the section 4.

\begin{figure*}[t]
\includegraphics[width=\textwidth,height=0.4\textwidth]{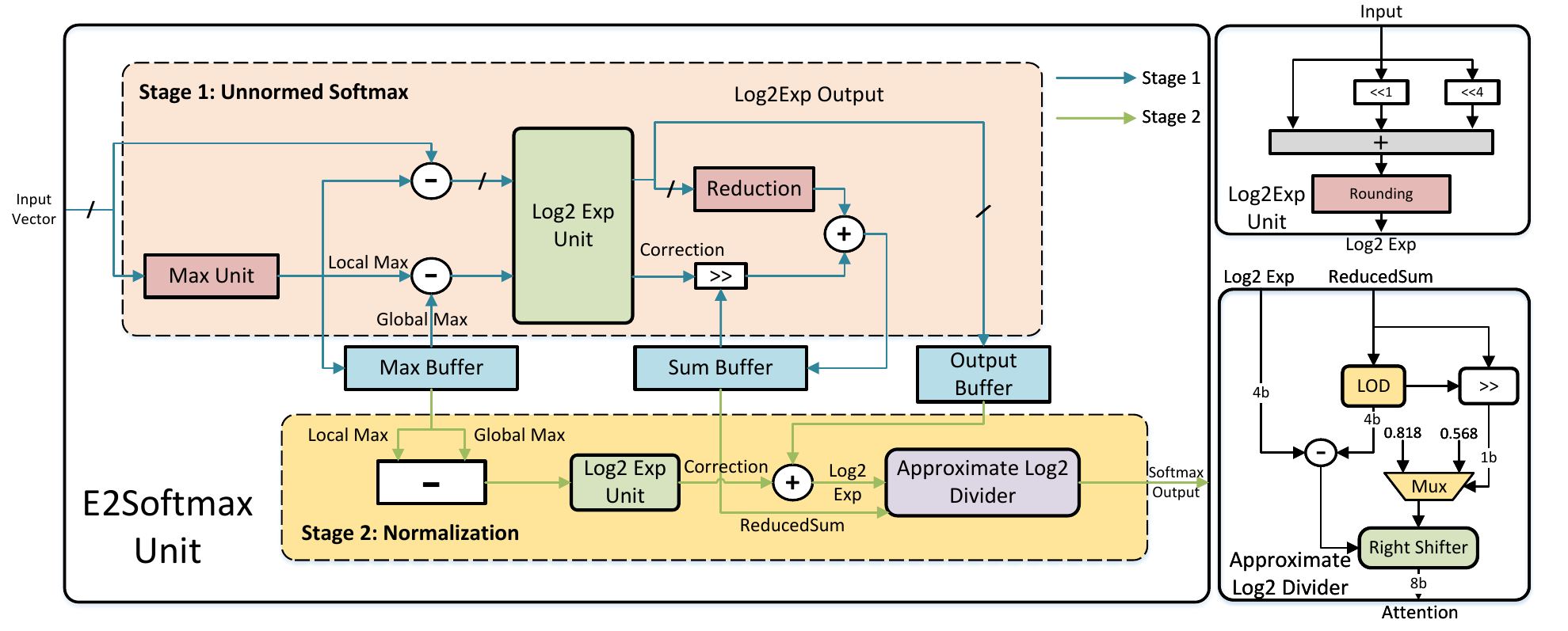}
  \centering
  \caption{Hardware design of E2Softmax Unit}
  \label{fig:5}
\end{figure*}
    \begin{algorithm}[H]
    \caption{Efficient Log2 Quantized Softmax}
    \label{E2S}
    \begin{algorithmic}[1]
    \State \textbf{Input:} $X \in \mathbb{R}^{L}$: Input Activation
    \State \textbf{Ouput:} $Y \in \mathbb{R}^L$: Softmax Output
            \State  $m_0 \gets -\infty$
            \State  $Sum \gets 0$
            \For{$i \gets $$ 1$ to $L$}             \Comment{Stage1}
                \State $m_i \gets$ Max$(X_i,m_{i-1})$
                \State $Y_i \gets$ Log2Exp$(X_i - m_i)$           
                \State $Sub \gets$ Log2Exp$(m_{i-1} - m_{i})$            
                \State $Sum \gets Sum >> Sub + 2^{-Y_i}$          
            \EndFor
            \For{$i$ $\gets$ $1$ to $L$}            \Comment{Stage2}
                \State $Sub \gets$ Log2Exp$(m_i - m_L)$
                \State $Y_i \gets$ ALDivision$(Sub+Y_i,Sum)$
            \EndFor
            \State \Return $Y$
    \end{algorithmic}
    \end{algorithm}
\subsection{AILayerNorm}
AILayerNorm algorithm utilizes dynamic compression to achieve low-precision statistic calculation based on the PTF quantization. The procedure of AILayerNorm is presented in Algorithm \ref{AIL}.

\noindent \textbf{Dynamic Compression.} LayerNorm utilizes mean and variance to normalize inputs. Therefore statistic calculation is a crucial part of the algorithm. To obtain the mean and variance, the expected value of the input and squared input ($E(x)$ and $E(x^2)$) are typically computed \cite{hw_ln}. While $E(x)$ can be computed using only addition, $E(x^2)$ requires a large number of multiplications. Therefore, we propose dynamic compression for low-precision statistic calculations, driven by the fact that small values are less important in the reduction of $x^2$ than in $x$:
\begin{equation}
    \frac{x_1^2}{x_1^2+x_2^2} < \frac{x_1}{x_1+x_2},\quad x_1 < x_2\text{ and } x_1,x_2 > 0
\end{equation}

\begin{figure*}[!t]
\includegraphics[width=\textwidth,height=0.3\textwidth]{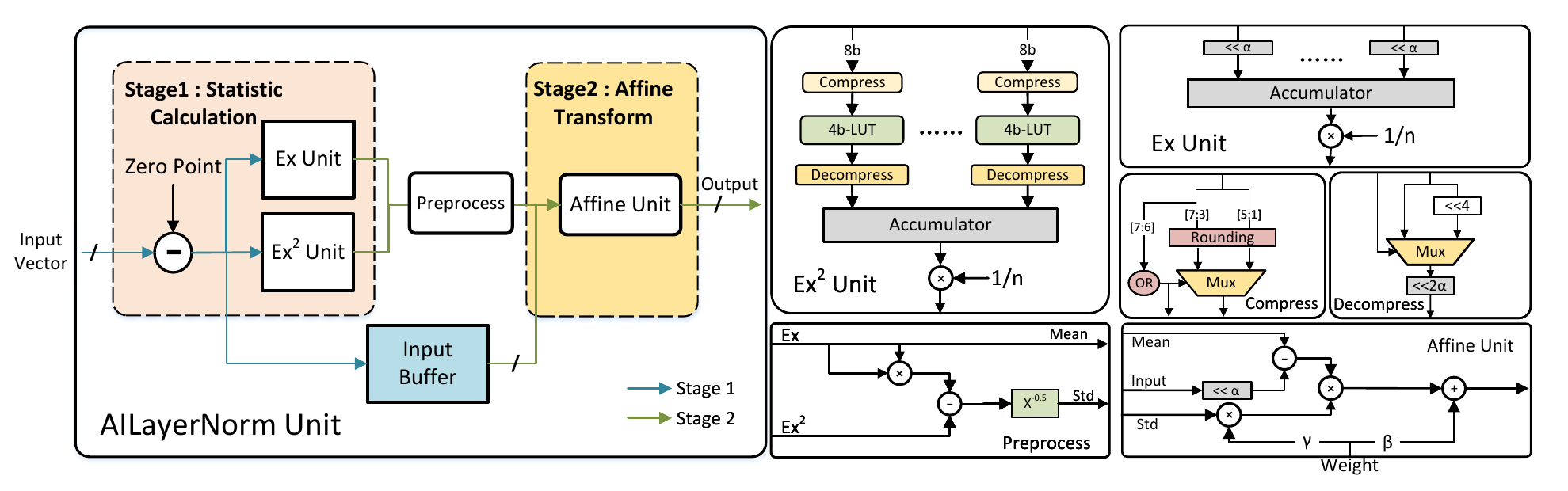}
  \centering
  \caption{Hardware design of AILayerNorm Unit}
  \label{fig:6}
\end{figure*}

To reduce computational complexity and minimize performance loss,  we propose a dynamic compression method for 8-bit unsigned integer $x[7:0]$ to generate a 4-bit approximation $y[3:0]$, which dynamically compresses inputs based on their value. As shown in Fig. \ref{fig:6}, we filter some less significant bits for compression. The length of bit-filtered is dynamically changed based on the value of input, which can be defined as $y,s = \text{Dynamic Compress}(x)$:

\begin{equation}
\resizebox{1.\hsize}{!}{$
\text{Dynamic Compress}(x) = 
    \begin{cases}
	\text{Clip}(\lfloor x / 2^4 \rceil,0,15),\ 1,\ x[7:6] \neq 0 \\
    \text{Clip}(\lfloor x / 2^2 \rceil,0,15),\ 0,\ \text{else}
    \end{cases}
    $}
\end{equation}

With our compression method, the calculation of $x^2$ can be done using 4-bit integer arithmetic and shift operation. A 1-bit signal $s$ is also generated to determine the length of bit shift (2 or 4) when recovering correct inputs. Experiments show that our method only induces errors of $0.2\%$ over $E(x^2)$  and $0.4\%$ over standard deviation with uniformly distributed input data.

\noindent \textbf{Low-precision statistic calculation.}  Our method is based on the PTF for LayerNorm quantization which addresses inter-channel variation of LayerNorm and compresses input acvtivation to 8-bit integer. However, channel-wise shift operations are required before statistic calculation to obtain accurate value of inputs, leading to high-precision calculation of variance. Specifically, 12-bit multiplication must be performed to calculate $X^2$ after bit-shifting of 8-bit integer $X$. Hence, We optimize the dataflow to tackle this problem based on an equivalent mathematical transformation. 
\begin{equation}
\begin{split}
    \text{Square}(X,\alpha) &= (X\textless \textless \alpha) \cdot (X\textless \textless \alpha) \\
     &= (X\cdot X) \textless \textless (2\alpha)
\end{split}
\end{equation}

It is arranged in the Decompress phase along with the decoding of dynamic compression. In conjunction with dynamic compression mentioned before, this optimization avoids 12-bit multiplication in statistic calculation and uses 4-bit integer arithmetic instead, contributing to low-precision statistic calculation.
    \begin{algorithm}[H]
    \caption{Approximate Integer Layer Normalization}
    \label{AIL}
    \begin{algorithmic}[1]
        \State \textbf{Input:} $X,\alpha:$ quantized input activation, power of two factor
        \State $\gamma,\beta,zp:$ affine weight, affine bias, zero point
        \State \textbf{Output: } $Y:$ quantized output activation
        \For{$i \gets$ $0$ to $C$} \Comment{Stage1}
            \State $X_{i} \gets$ $X_i - zp$
            \State $X_{c},s \gets$ Dynamic Compress($X_{i}$) \Comment{Compress}
            \State $X_{c} \gets$ $X_{c}^{2} \textless\textless (4s + 4)$ \Comment{Square \& Decompress}
            \State $E_x \gets E_x + X_{i} \textless \textless  \alpha_{i}$             \Comment{PTF Shift}
            \State $E_{x^2} \gets$ $E_{x^2} + X_{c} \textless \textless  (2\alpha_{i})$
        \EndFor
        \State $E_x,E_{x^2} \gets$ $E_x \cdot \frac{1}{C}, (E_{x^2} \textless \textless  4) \cdot \frac{1}{C}$
        \State $\mu,std_{inv} \gets$ $E_x$, $(E_{x^2} - (E_x)^2)^{-\frac{1}{2}}$
        \For{$i \gets$ $0$ to $C$} \Comment{Stage2}
            \State $A , B \gets$ $\gamma_i \cdot std_{inv}$ , $\beta_i$
            \State $X_i \gets$ $(X_i \textless \textless  \alpha_i - \mu)$
            \State $Y_i \gets$ $A\cdot X_i + B$
        \EndFor       
        \State \Return $Y$
    \end{algorithmic}
    \end{algorithm}
    
\section{SOLE Hardware}
In SOLE, we design custom hardware units as commodity hardware platforms do not support the special function units involved in algorithms. With custom hardware design, it can be integrated into other dedicated accelerators or tensor processing hardware like GPU tensor cores to accelerate transformer inference.

\subsection{E2Softmax Unit}
We propose \textrm{E2Softmax Unit} to implement our algorithm. Its key features include the Log2Exp Unit and Approximate Log-based Divider, which are implemented in a LUT-free and multiplication-free manner to achieve efficiency. As shown in Fig. \ref{fig:5}, the computation process is divided into two stages to organize the dataflow. Buffers are designed ping-pong to support pipeline.

\noindent \textbf{Unnormed Softmax.} Stage 1 consists of three subunits, namely Max Unit, Log2Exp Unit and Reduction Unit. It receives a slice of the input vector since the length of vector can be as large as 1024. The Max Unit determines the local maximum of input vector by using a comparison tree. Then it is subtracted from the input vector and $Global Max$ read from buffer in case $LocalMax$ is larger than $GlobalMax$. The Log2Exp Unit receives them as inputs to generate outputs by shifts and additions, while modern hardware usually requires LUTs and multiplier. Rounding operations are applied at the end to get 4-bit integer outputs, namely $Log2Exp\ Output$ and $Correction$. The $Log2Exp\ Output$ are stored in the Output Buffer for Stage 2 and sent to the Reduction Unit at the same time, while the $Correction$ modifies the data from Sum Buffer for online normalization. Since $Log2Exp\ Output$ are quantized to 4-bit, little memory is required to store the intermediate result.

\noindent \textbf{Normalization.} 
Once the computation of reduced sum is completed, Stage 2 starts to perform division and generate softmax output. While division is generally approximated by linear piece-wise functions that rely LUTs and high-precision multipliers in prior works\cite{Softermax,NN-LUT}, we introduce the Approximate Log-based Divider in our design. $Correction$ is firstly computed and added to generate $Log2Exp$ as modification, then $Log2Exp$ and $ReducedSum$ are sent to divider to obtain outputs. The Approximate Log-based Divider is simple and light-weight, consisting of a Leading-one detector (LOD), a subtractor, a two-way multiplexer and two shifters. As mentioned in algorithm \ref{E2S}, the characteristic part of $ReducedSum$ is produced by LOD and subtracts $Log2Exp$. Then it Selects the bit next to the leading-one bit by shifting and generate a 1-bit result. It is functioned as select signal $s$ of the Multiplexer to decide the output, which equals with the formula that:
\begin{equation}
        O = \frac{1.636-0.5s}{2} =  \left\{
	\begin{aligned}
	0.818&,\quad s = 0 \\
    0.568&,\quad s = 1
	\end{aligned}
	\right.
\end{equation}

The result is then scaled using right shifter with respect to the subtraction result, thus obtaining the final output of E2Softmax. Compared with previous work, our divider utilizes the logarithm input produced by 
\textrm{Log2Exp} Unit, involving only a subtraction, a LOD, a two-way multiplex and two shift operations.

\subsection{AILayerNorm Unit}
We design AILayerNorm Unit characterized by low-precision statistic calculation to implement AILayerNorm algorithm. It is also divided into two stages to handle statistic calculation and affine transform. Ping-pong buffers are utilized to design a pipelined unit. Unlike the batch normalization, the layer normalization does not impose any restriction on the size of a mini-batch. So it is able to be used in the pure online regime with the batch size equal to 1 \cite{layernorm}.

\noindent \textbf{Statistic Calculation.} Layer normalization computes mean and variance for normalization of inputs. Stage 1 is in charge of the statistic calculation. It consists of two subunits, namely Ex Unit and $\text{Ex}^2$ Unit. While prior works require high-precision calculation for mean and variance which introduce power and area inefficiency, for example INT32 for I-BERT \cite{I-BERT}, our design is capable of doing so in a low-precision manner. The 8-bit quantized inputs firstly subtract zero point and the results are sent to Ex, $\text{Ex}^2$ Unit and Input Buffer. In Ex Unit, inputs are scaled with power of two factors and then gathered to perform 12-bit reduction. In $\text{Ex}^2$ Unit, dynamic compression are firstly applied to obtain 4-bit inputs. Since square function with 4-bit inputs only have 16 possible outcomes, we implement square function with 16-entry Look-up table instead of multiplier to achieve better efficiency. Scaling operations are performed in the Decompress Unit as compensation for PTF and Dynamic Compression. Accumulator receives the outputs to perform reduction. After $\sum_{i=1}^{n}X_i$ and $\sum_{i=1}^{n} X_i^2$ are obtained, other operations such as power of -0.5 ($x^{-0.5}$) function are used in Preprocess Unit to generate parameters that Stage 2 requires. The $x^{-0.5}$ Unit is implemented using a LUT in our design which takes up little area and power consumption due to its small operation density.

\noindent \textbf{Affine Transform.} After Stage 1 finishes the calculation of statistic, Stage 2 starts to normalize inputs and rescale them with affine parameters $\gamma$, $\beta$. Catering to Algorithm \ref{AIL}, the Affine Unit fuses normalization and rescale operation into two multiplication and two addition. The first multiplier computes $A$ with weights and $Std$. Inputs read from Input Buffer are scaled with PTF and subtracted with $\mu$ from $Ex$ Unit. At last, multiplication and accumulation are performed to obtain outputs $Y = A\cdot X + B$. It should be noted that weights are also quantized to 8-bit integers.

\section{Evaluation}
\subsection{Experiment Setting}
\noindent \textbf{Software setup.}
To validate our algorithm, we conducted exclusive experiments on CV and NLP tasks. For CV tasks, we selected ImageNet-1K\cite{imagenet} as our benchmark with different vision transformers, i.e., DeiT\cite{DeiT}, Swin Transformer\cite{Swin}. For NLP tasks, we conducted experiments on the GLUE benchmark\cite{glue} and SQuAD v1.1 dataset\cite{squad} with BERT-Base model \cite{BERT}. 
FP32 and \text{INT8} models were selected as our baseline. We chose \text{INT8} model as baseline to demonstrate that SOLE can serve as a plugin for other compression method like quantization. We further applied SOLE algorithms as a replacement for the floating-point version to evaluate the impact of SOLE on accuracy, which was noted as \text{FP32+SOLE} and \text{INT8+SOLE}. INT8 models in Table \ref{table1} are obtained through post-training quantization following the settings in \cite{FQ-VIT} while in Table \ref{table2} we performed quantization-aware fine-tuning with 8-bit weights and activation to get INT8 models \cite{int_q}.
\begin{table}[!t]
	\centering  
	\caption{Algorithm Experiment on ImageNet-1K benchmark.}  
	\label{table1}  
    \renewcommand\arraystretch{1.5}{
    \resizebox{0.9\linewidth}{0.12\linewidth}{
	\begin{tabular}{|c|c|c|c|c|c|c|c|}  
		\hline  
		Model&DeiT-T&DeiT-S&DeiT-B&Swin-T&Swin-S&Swin-B \\
		\hline
        FP32& 72.21 & 79.85 & 81.85 &  81.38 & 83.23 & 83.60 \\ 
        \hline
        FP32 + SOLE&        $\mathbf{71.33}$ & $\mathbf{79.27}$ & $\mathbf{81.60}$ & $\mathbf{80.58}$ & $\mathbf{82.75}$ & $\mathbf{83.05}$ \\
        \hline
        INT8          & 71.72  & 79.25 & 81.42 & 80.42 & 82.84  & 83.14 \\
		\hline
        INT8 + SOLE   & $\mathbf{71.07}$ & $\mathbf{78.89}$ & $\mathbf{81.12}$ &  $\mathbf{80.14}$ & $\mathbf{82.60}$ & $\mathbf{82.79}$ \\
        \hline
	\end{tabular}
 }
 }
\end{table}

\noindent \textbf{Hardware setup.}
To evaluate the impact of SOLE on speedup and hardware efficency, we implemented SOLE hardware in RTL-Verilog and synthesized RTL using Synopsys Design Compiler on 28nm TSMC library with target 1ns clock period (1GHz). Power consumption was estimated at the typical corner by PrimeTimePX, with the switching activity from VCD simulation traces. Vector size of SOLE hardware was set as 32 to match the MAC throughput of traditional accelerators. 

For speedup evaluation, we chose NVIDIA 2080Ti GPU as our baseline. We scaled up our E2Softmax Unit and AILayerNorm unit resource by 32 times to make relatively fair comparisons. For area and power evaluation, apart from GPU, we compare SOLE with state-of-the-art custom hardware. Softermax \cite{Softermax} was selected as the baseline for Softmax while NN-LUT \cite{NN-LUT} was chosen as the baseline for LayerNorm \footnote{In NN-LUT, the implement of non-linear operations adopted NN-based LUT design while the other followed the method of I-BERT.}. We re-implemented these designs under the same setting with SOLE to extract power and area for fair comparison.

\subsection{Algorithm Performance}
Table \ref{table1} and Table \ref{table2} summarize the accuracy comparison on different transformer models, across CV and NLP tasks. Firstly, we compare the accuracy of \text{FP32} and \text{FP32+SOLE}. The results show that SOLE incurs negligible accuracy drops. The worst accuracy drop is under \textrm{0.9\%} while the average accuracy drop is nearly \textrm{0.38\%} on different datasets and models. Secondly, we compare the performance of \text{INT8} and \text{INT8+SOLE}. 
The results reveal that SOLE functions well in cooperation with INT8 quantization with the worst accuracy drop at \textrm{0.8\%} and the average drop at \textrm{0.2\%}, demonstrating that SOLE can be integrated with other model compression methods. 
Note that only matrix multiplication is done under INT8 format and other operations including Softmax and LayerNorm are still under FP32 format in \text{INT8} models. Through integration with SOLE, Softmax and LayerNorm can be calculated with the input and output in 8-bit format.
Another notable advantage of SOLE is that SOLE maintains inference accuracy without additional training or fine-tuning, therefore avoiding expensive training overhead and making it convenient to use.

\begin{table}[!t]\normalsize
	\centering  
	\caption{Algorithm Experiment on SQuAD and GLUE tasks with BERT-Base model.}  
	\label{table2}  
    \renewcommand\arraystretch{1.5}{
    \resizebox{1.0\linewidth}{0.13\linewidth}{
	\begin{tabular}{|c|c|c|c|c|c|c|c|c|}  
		\hline
		Benchmark& CoLA & MRPC & SST-2 & QQP & MNLI & QNLI & RTE  & SQuAD\\
        \hline
        FP32& 83.49 & 85.66 & 92.19 &  91.35 & 84.06 & 91.63 & 62.81 &88.17 \\ 
        \hline
        FP32 + SOLE &        $\mathbf{83.56}$ & $\mathbf{86.38}$ & $\mathbf{91.52}$ & $\mathbf{91.00}$ & $\mathbf{84.09}$ & $\mathbf{90.91}$ & $\mathbf{62.62}$ &$\mathbf{87.57}$\\
        \hline
        INT8          & 82.25  & 86.62 & 91.96 & 90.71 & 83.13  & 89.11 & 63.49 & 87.16\\
		\hline
        INT8 + SOLE   & $\mathbf{82.72}$ & $\mathbf{85.74}$ & $\mathbf{91.41}$ &  $\mathbf{90.39}$ & $\mathbf{82.89}$ & $\mathbf{88.70}$ &  $\mathbf{65.39}$ & $\mathbf{86.45}$\\
        \hline
	\end{tabular}
 }
 }
\end{table}
\subsection{Speedup}
\begin{figure}[!t]
\includegraphics[width=0.5\textwidth,height=0.42\textwidth]{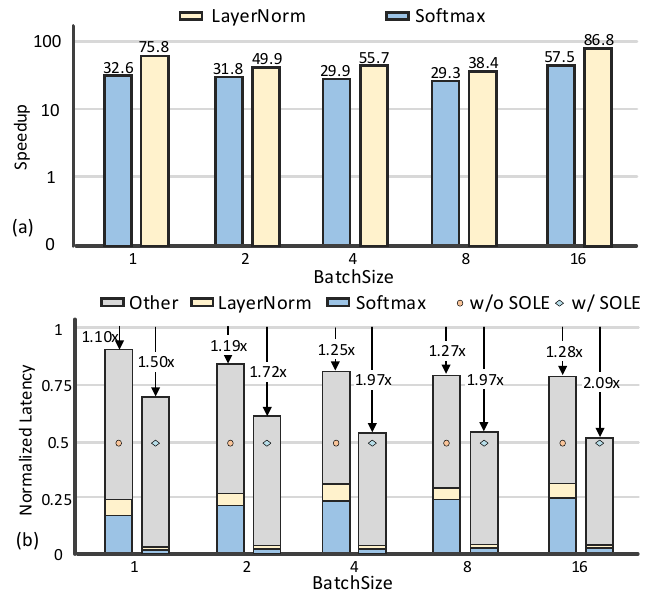}
  \caption{(a) Speedup over GPU on Softmax and LayerNorm. (b) End-to-end speedup and latency breakdown, the latency is normalized with respect to FP32 implementation.}
  \label{fig:7}
\end{figure}

Fig. \ref{fig:7} reveals the speedup of SOLE over 2080Ti GPU. We evaluate both stand along Softmax/LayerNorm operations on DeiT-Tiny, across different batchsize from 1 to 16. Token length is set as 785 corresponding to $448\times448$ image size. We also test end-to-end speedup of INT8 model over FP32 model with and without our HW/SW optimization.

As illustrated in Fig. \ref{fig:7}(a), compared with GPU, SOLE achieves 29.3$\times$-57.5$\times$ and 38.4$\times$-86.8$\times$ speedup on Softmax and LayerNorm with average speedup at 36.2$\times$ and 61.3$\times$, respectively. The speedup mainly comes from that we design highly specialized and pipelined datapath to reduce latency. In SOLE, ping-pong buffers and online normalization are utilized to maximize throughput. 
In Fig. \ref{fig:7}(b), we can observe that INT8 model only achieve 1.10$\times$ to 1.28$\times$ speedup over FP32 model in GPU. It is due to the fact that INT8 quantization only alleviates overhead in matrix multiplication whereas non-linear operations like Softmax and LayerNorm occupy a significant portion of the inference. However, with the optimization of SOLE in non-linear operations, the burden can be significantly alleviated, resulting in 1.50$\times$ to 2.09$\times$ end-to-end speedup over FP32 model.

It has been demonstrated in section 5.2 that algorithms of SOLE can be integrated with INT8 quantization with negligible accuracy drop. The hardware of SOLE can also be easily integrated into modern accelerators since its 8-bit input and output are compatible with existing 8-bit integer vector MAC datapaths in accelerators and modern GPU tensor cores. Thus, comprehensive optimization can be achieved to get higher speedup. It is worth noticing that we simply use 32 E2Softmax Units and AILayerNorm Units to compare with GPU whose hardware resource still dominates ours. Whereas, SOLE offers scalable acceleration over Softmax, LayerNorm and end-to-end inference for transformer.
\subsection{Hardware efficiency}
\noindent \textbf{Energy-efficiency.}
We first compare energy-efficiency of SOLE with Softermax, NN-LUT and 2080Ti GPU in Table \ref{table3}. The setting is the same as mentioned in section 5.3 and we report the average energy-efficiency in Table \ref{table3}.For in-depth evaluation, we compare the power consumption of the subunits, i.e. \textit{Normalization Unit} and \textit{Statistic Unit} mentioned in Section 4, which reveal computation power overhead. We also examine the complete hardware units, i.e. \textit{Softmax Unit} and \textit{LayerNorm Unit}, to measure both computation and memory overhead, thereby providing a comprehensive measure of power consumption. We find that SOLE offers 2.46$\times$ and 11.3$\times$ energy-efficiency improvements in \textit{Normalization Unit} and \textit{Statistic Unit}, resulting in hardware that are 3.04$\times$ and 3.86$\times$ more energy efficient for Softmax and LayerNorm respectively. The difference in energy-efficiency between subunits and complete hardware stems from the fact that power consumption in complete hardware is mainly determined by memory overhead, as discussed in Section 1. Overall , our design averagely deliver 3.04$\times$, 4925$\times$ energy-efficiency improvements over Softermax and GPU in Softmax and 3.86$\times$, 4259$\times$ energy-efficiency improvements over NN-LUT and GPU in LayerNorm. 

Results have shown prominent energy-efficiency improvements in both computation and memory. For Softmax, the computational advantage is due to the use of the Log2Exp Unit and Approximate Log-based Divider in E2Softmax Unit. These units utilize shift and addition to implement non-linear operations while prior works need LUTs and multiplication. Additionally, log2 quantization on the exponent output allows intermediate results to be temporarily stored in 4-bits, greatly reducing the cost of memory access compared to the 16-bits required by Softermax. For LayerNorm, we adopt dynamic compression and low-precision statistic calculation in AILayerNorm to avoid INT32 multiplication in NN-LUT entirely, instead using only 16-entry LUTs and shift operation, resulting in significant energy savings for computation. Futhermore, intermediate memory access is saved to a large extent as SOLE quantizes the input data to 8-bit while prior works need to store 32-bit data. In general, SOLE significantly outperforms prior works in energy-efficiency.

\noindent \textbf{Area-efficiency}
As shown in Table \ref{table3}, SOLE also exhibits enhancements in terms of area compared with the state-of-the-art designs. Specifically, SOLE achieves 2.89$\times$ and 3.79$\times$ area-efficiency improvements in \textit{Normalization Unit} and \textit{Statistic Unit} while offer 2.82$\times$ and 3.32$\times$ area-efficiency improvements for complete \textit{Softmax Unit} and \textit{LayerNorm Unit} at the same time. The area saving in computation subunits comes from the simplified implementation. In Softmax, SOLE only needs shifters and adders while Softermax relies on multipliers and LUTs. Similarly, SOLE utilizes 16-entry LUTs and shifters to replace 32-bit multipliers in LayerNorm. The entire calculation process is based on 8-bit input data so that the computational area burden is alleviated. Another aspect of the area-efficiency directly comes from the decreased size of buffer since SOLE reduces the bit-width of data stored in buffer from 16/32-bit to 4/8-bit, for Softmax and LayeNorm respectively. 
\begin{table}[!t]
	\centering
	\caption{SOLE Comparison to Softermax, NN-LUT and GPU in energy and area.}
    \label{table3}
    \resizebox{0.5\textwidth}{!}{
	\begin{tabular}{cccc}
		\toprule  
		 Baseline& &Energy-Efficiency&Area-Efficiency \\ 
		\midrule  
		\multirow{2}{*}{Softermax\cite{Softermax}}& Normalization Unit&2.46$\times$&2.89$\times$\\
        \cmidrule(r){2-4} 
        & Softmax Unit  &3.04$\times$ & 2.82$\times$\\
        \midrule
         \multirow{2}{*}{NN-LUT\cite{NN-LUT}} &Statistic Unit & 11.3$\times$ & 3.79$\times$\\
        \cmidrule(r){2-4}
        & LayerNorm Unit &3.86$\times$ & 3.32$\times$\\
        \midrule
        \multirow{2}{*}{2080Ti GPU} & Softmax Unit & 4925$\times$ & - \\
        \cmidrule(r){2-4}
        & LayerNorm Unit & 4259$\times$ & - \\
		\bottomrule  
	\end{tabular}
    }
\end{table}

\section{Conclusion}
We propose SOLE, a hardware/software co-design method to enable efficient Softmax and LayerNorm inference in transformer. SOLE consists of E2Softmax and AILayerNorm which implement Softmax and LayerNorm with low-precision calculation and low bit-width data storage to achieve better efficiency. Experiments show that SOLE incurs negligible accuracy drop without additional training and can be integrated with orthogonal compression method like quantization. SOLE achieves orders of magnitude speedup and energy savings over GPU while offering 3.04$\times$, 3.86$\times$ energy-efficiency improvements and 2.82$\times$, 3.32$\times$ area-efficiency improvements over Softermax and NN-LUT, respectively.

{\footnotesize
    \bibliographystyle{ieeetr}
    \bibliography{egbib}
}

\vspace{12pt}

\end{document}